\documentclass[letterpaper, 10 pt, conference]{ieeeconf}  

\IEEEoverridecommandlockouts                              

\usepackage{url}
\usepackage{amsmath}
\usepackage{graphicx}
\usepackage{verbatim}
\usepackage{algorithm}
\usepackage{algorithmic}
\usepackage{hyperref}
\usepackage{color}
\usepackage{wrapfig}
\usepackage{xcolor}
\usepackage{tikz}
\usetikzlibrary{shapes,arrows}
\usetikzlibrary{positioning,fit,calc}
\usepackage{setspace}
\usepackage{calc}
\usepackage[small]{caption}
\usepackage{balance}

\long\def\ignorethis#1{}

\newcommand{\tr}{^\mathrm{T}}

\newcommand{\gauss}{\mathcal{N}}

\newcommand{\vnorm}[1]{\|#1\|}

\newcommand{\channel}{c}
\newcommand{\softmax}{\mathbf{s}}
\newcommand{\softmaxpix}{s_{cij}}

\newcommand{\responsepix}{a_{cij}}
\newcommand{\responsepixprime}{a_{ci'j'}}

\newcommand{\trajdist}{p}

\newcommand{\cost}{\ell}
\newcommand{\state}{\mathbf{x}}
\newcommand{\action}{\mathbf{u}}

\newcommand{\tst}{\tilde{\state}_t}

\newcommand{\traj}{\tau}

\newcommand{\ucovart}{\mathbf{C}_t}

\newcommand{\kl}{D_\text{KL}}

\newcommand{\fct}{f_{c t}}
\newcommand{\fxt}{f_{\state t}}
\newcommand{\fut}{f_{\action t}}

\newcommand{\kpol}{\mathbf{k}}
\newcommand{\Kpol}{\mathbf{K}}

\newcommand{\samples}{\mathcal{S}}

\newcommand{\noise}{\mathbf{F}}

\newcommand{\costgradt}{\cost_{\state\action t}}
\newcommand{\costhesst}{\cost_{\state\action,\state\action t}}

\newcommand{\st}{\state_t}
\newcommand{\at}{\action_t}

\newcommand{\img}{I}
\newcommand{\feat}{\mathbf{f}}
\newcommand{\autoenc}{h_\text{enc}}   
\newcommand{\autodec}{h_\text{dec}} 
\newcommand{\temp}{\alpha} 
\newcommand{\featidx}{\mathcal{I}}
\newcommand{\featidxsel}{{\mathcal{I}_s}}

\newcommand{\specialcell}[2][c]{%
  \begin{tabular}[#1]{@{}c@{}}#2\end{tabular}}

\title{\LARGE \bf
Deep Spatial Autoencoders for Visuomotor Learning
}

\author{Chelsea Finn, Xin Yu Tan, Yan Duan, Trevor Darrell, Sergey Levine, Pieter Abbeel
\thanks{Department of Electrical Engineering and Computer Science, University of California, Berkeley, Berkeley, CA 94709}%
}

\begin{document}

\maketitle
\thispagestyle{empty}
\pagestyle{empty}

\begin{abstract}

Reinforcement learning provides a powerful and flexible framework for automated acquisition of robotic motion skills. However, applying reinforcement learning requires a sufficiently detailed representation of the state, including the configuration of task-relevant objects. We present an approach that automates state-space construction by learning a state representation directly from camera images. Our method uses a deep spatial autoencoder to acquire a set of feature points that describe the environment for the current task, such as the positions of objects, and then learns a motion skill with these feature points using an efficient reinforcement learning method based on local linear models. The resulting controller reacts continuously to the learned feature points, allowing the robot to dynamically manipulate objects in the world with closed-loop control. We demonstrate our method with a PR2 robot on tasks that include pushing a free-standing toy block, picking up a bag of rice using a spatula, and hanging a loop of rope on a hook at various positions. In each task, our method automatically learns to track task-relevant objects and manipulate their configuration with the robot's arm.

\end{abstract}

\section{Introduction}

One of the fundamental challenges in applying reinforcement learning to robotic manipulation tasks is the need to define a suitable state space representation. Typically, this is handled manually, by enumerating the objects in the scene, designing perception algorithms to detect them, and feeding high-level information about each object to the algorithm. However, this highly manual process makes it difficult to apply the same reinforcement learning algorithm to a wide range of manipulation tasks in complex, unstructured environments. What if a robot could automatically identify the visual features that might be relevant for a particular task, and then learn a controller that accounts for those features? This would amount to automatically acquiring a vision system that is suitable for the current task, and would allow a range of object interaction skills to be learned with minimal human supervision. The robot would only need to see a glimpse of what task completion looks like, and could then figure out on its own how to change the scene into that configuration.

Directly learning a state space representation from raw sensory signals, such as images from a camera, is an active area of research \cite{rlv-arlrv-12,jb-srl-14}, and while considerable progress has been made in recent years \cite{wsbr-e2c-15,lr-avsrg-13}, applications to real robotic systems remain exceedingly difficult. The difficulties stem from two challenges. First, learning a good representation with unsupervised learning methods, such as deep learning, often requires a very large amount of data \cite{lbh-dl-15}. Second, arbitrary learned visual representations do not always lend themselves well to control. We address these challenges by using a spatial autoencoder architecture to learn a state representation that consists of feature points. Intuitively, these feature points encode the configurations of objects in the scene, and the spatial autoencoder that we describe provides for data-efficient learning by minimizing the number of non-convolutional parameters in the encoder network. Since our learned feature points correspond to object coordinates, this architecture also addresses the second challenge, since real-valued quantities such as positions are more amenable to control than the types of discrete or sparse features that are more commonly learned with deep learning methods \cite{lca-sdbnm-08}.

In fact, our experiments show that our learned feature point representation can be used effectively in combination with an efficient trajectory-centric reinforcement learning algorithm. This method iteratively fits time-varying linear models to the states visited at the previous iteration \cite{la-lnnpg-14}, estimating how the robot's actions (which correspond to joint torques) affect the state. When the state includes visual feature points, this method can learn how the robot's actions affect the objects in the world, and the trained controllers can perform closed-loop control on the configuration of these objects, allowing the robot to move and manipulate them. Furthermore, this reinforcement learning algorithm can easily be combined with guided policy search to learn complex, nonlinear policies that generalize effectively to various initial states \cite{lwa-lnnpg-15}, which we demonstrate by training a nonlinear neural network policy on top of our learned representation.

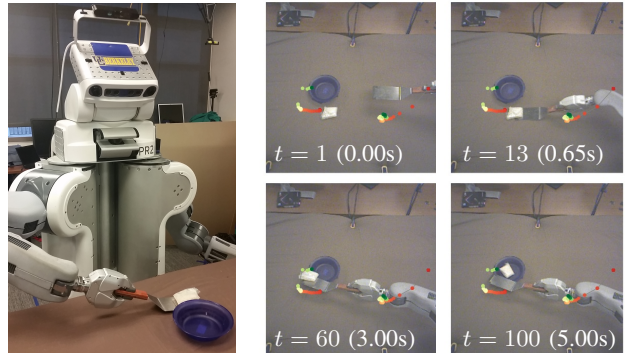
\begin{figure}
\setlength{\unitlength}{0.5\columnwidth}
\begin{picture}(2.0,1.07) \linethickness{0.5pt}

\put(0.04,-0.02){\includegraphics[width=0.35\columnwidth]{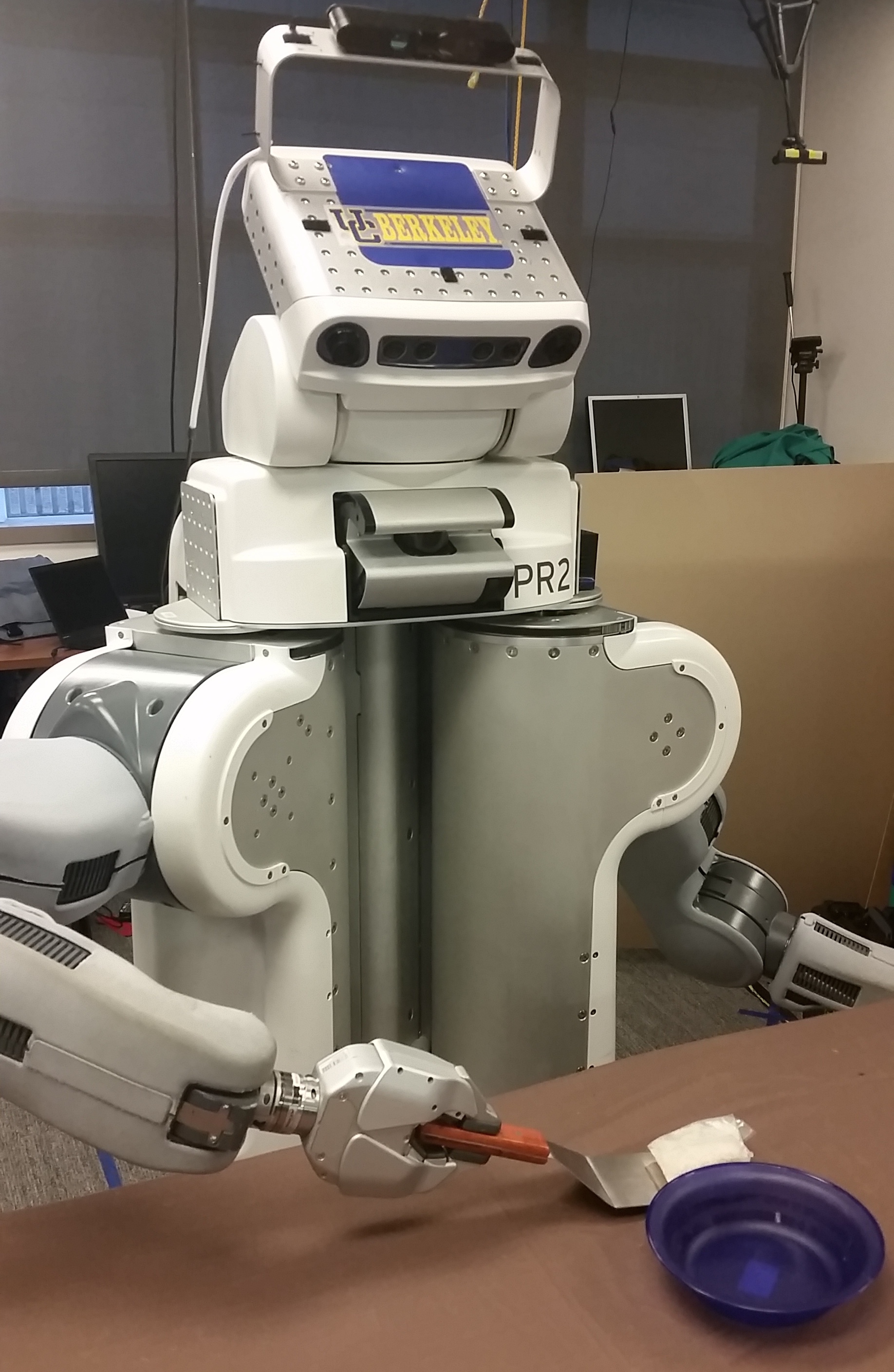}}

\put(0.79,0.5){\includegraphics[width=0.29\columnwidth]{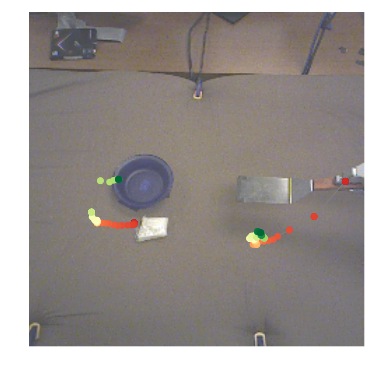}}
\put(1.36,0.5){\includegraphics[width=0.29\columnwidth]{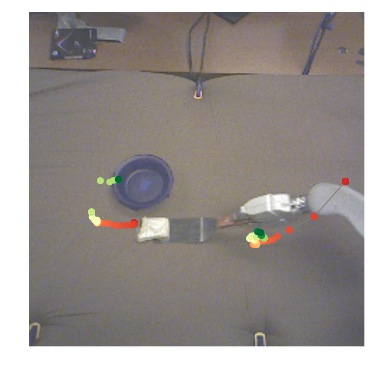}}
\put(0.79,-0.06){\includegraphics[width=0.29\columnwidth]{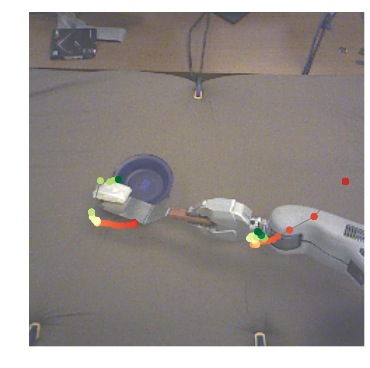}}
\put(1.36,-0.06){\includegraphics[width=0.29\columnwidth]{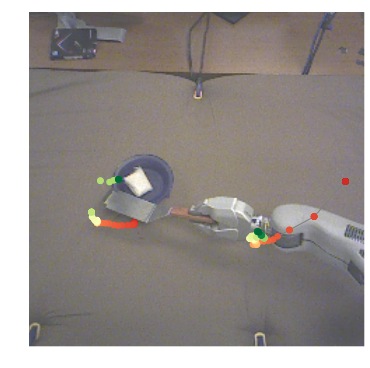}}

\put(0.86,0.57){\textcolor{white}{\small{$t=1$ (0.00s)}}}
\put(1.44,0.57){\textcolor{white}{\small{$t = 13$ (0.65s)}}}
\put(0.86,0.0){\textcolor{white}{\small{$t = 60$ (3.00s)}}}
\put(1.44,0.0){\textcolor{white}{\small{$t = 100$ (5.00s)}}}

\end{picture}
\caption{PR2 learning to scoop a bag of rice into a bowl with a spatula (left) using a learned visual state representation (right).
\label{fig:teaser}
}
\vspace{-0.2in}

\end{figure}

Our main contribution is a method for learning vision-based manipulation skills by combining unsupervised learning using deep spatial autoencoders with simple, sample-efficient trajectory-centric reinforcement learning. We demonstrate that this approach can be used to learn a variety of manipulation skills that require ``hand-eye'' coordination, including pushing a free-standing toy block, scooping objects into a bowl, using a spatula to lift a bag of rice from the table (shown in Figure~\ref{fig:teaser}), and hanging a loop of rope on hooks at various positions. Each of these tasks are learned using the same algorithm, with no prior knowledge about the objects in the scene, and using only the onboard sensors of the PR2 robot, which consist of joint encoders and an RGB camera. We also compare our approach to a number of baselines that are representative of previously proposed visual state space learning methods.

\section{Related Work}

Reinforcement learning and policy search methods have been applied to perform a variety of robotic tasks, such as table tennis \cite{kop-rlarm-10}, object manipulation \cite{drf-lclcm-11,phas-lgmsl-09}, and locomotion \cite{emmnc-lcbbl-08,gpw-fbwrc-06,kp-pgrlf-04,tzs-spgrl-04}. One key challenge to using RL successfully is handling high-dimensional sensory observations needed to perform sensorimotor control. A standard approach is to hand-design low-dimensional features from the observations; however, this requires manual task-specific engineering, which undermines the goal of using an automatic learning-based approach. Other approaches instead use function approximation to replace the value function in standard RL algorithms like TD-learning \cite{sh-rlfsa-04} and approximate Q-learning \cite{b-rlwfa-95}. These approaches typically require large amounts of time and data to learn the parameter space. In this work, we use the approach of first learning a low-dimensional state representation from observations via unsupervised learning, and then use a data-efficient RL algorithm with the learned state representation to learn robotic manipulation tasks involving perception.

Prior work has proposed to learn representations using general priors from the physical world~\cite{jb-srl-14} or the ability to predict future observations~\cite{sljps-lpsr-03,bsg-psr-11,bsbro-alsr-15}. Representation learning has been applied to control from high-dimensional visual input in several recent works. Lange et al.~\cite{rlv-arlrv-12} use autoencoders to acquire a state space for Q-learning and evaluate on simple real-world problems with 2 to 3 dimensions. Several methods also train autoencoders that can predict the next image, effectively learning dynamics~\cite{wsbr-e2c-15,gml-lluu-15}, though such methods have only been demonstrated on toy problems with synthetic images. The quantity of data required to apply deep learning methods to real-world images is known to far exceed that of toy tasks~\cite{lbh-dl-15}, and prior methods generally avoid this issue by using synthetic images. We also use an autoencoder, but we use a spatial architecture that allows us to acquire a representation from real-world images that is particularly well suited for high-dimensional continuous control. Our architecture is also data-efficient and can handle relatively small datasets. We demonstrate our method on a range of real-world robotic tasks.

Other methods have been proposed to train policies from high-dimensional observations, without first acquiring a low-dimensional state space \cite{mksga-padrl-13,levine2015end}. However, running these methods on a real robotic platform requires either impractical amounts of data \cite{mksga-padrl-13} or an instrumented training setup that provides knowledge about relevant objects at training time \cite{levine2015end}. Our method does not require this knowledge, which enables it to learn tasks such as pushing, tossing, and scooping without using any additional instrumentation. In fact, our method has no knowledge of which objects make up the scene at all. These tasks cannot be learned using our prior method \cite{levine2015end} without mechanisms such as motion capture.

The goal of our method is similar to that of visual servoing, which performs feedback control on image features \cite{ecr-navsr-92,mkd-vbcqp-14,whb-reecu-96}. However, our controllers and visual features are learned entirely from real-world data, rather than being hand-specified. This gives our method considerable flexibility in how the visual input can be used and, as a result, the tasks that it can learn to perform. Furthermore, our approach does not require any sort of camera calibration, in contrast to many visual servoing methods (though not all -- see e.g. \cite{jfn-eeuvs-97,ya-auvs-94}).

Most deep learning architectures focus on semantic representation learning, e.g.~\cite{sljsr-gdwc-14}. Some spatial network architectures have been proposed for tasks such as video prediction and dynamics modeling~\cite{zmgl-wwae-15,gml-lluu-15}. The architecture that we use for the encoder in our unsupervised learning method is based on a spatial softmax followed by an expectation operation, which produces a set of spatial features that correspond to points of maximal activation in the last convolutional layer. We previously proposed this type of architecture in the context of policy search for visuomotor policies \cite{levine2015end}. However, in our prior work, this type of network was trained with supervised learning, regressing directly to joint torques. Obtaining these joint torque targets requires being able to solve the task, which requires a representation that is already sufficient for the task. In this paper, we show how the representation can be learned from scratch using unsupervised learning, which extends the applicability of this architecture to a much wider range of tasks where a suitable representation is not known in advance.

\section{Preliminaries}
\label{sec:rl}

In this paper, we are primarily concerned with the task of learning a state representation for reinforcement learning (RL) from camera images of robotic manipulation skills. However, we consider this problem in the context of a specific RL algorithm that is particularly well suited to operate on our learned state representations. We describe the RL method we use in this background section, while the Section~\ref{sec:ae} describes our representation learning method. The derivation in this section follows prior work \cite{la-lnnpg-14}, but is repeated here for completeness.
\vspace{-0.5cm}
\subsection{RL with Local, Time-Varying Linear Models}

Let $\st$ and $\at$ be the state and action at time step $t$. The actions correspond to motor torques, while states will be learned using the method in the next section. The aim of our RL method is to minimize the expectation $E_{\trajdist(\traj)}[\cost(\traj)]$ over trajectories $\traj = \{\state_1,\action_1,\dots,\state_T,\action_T\}$, where \mbox{$\cost(\traj) = \sum_{t=1}^T \cost(\st,\at)$} is the total cost, and the expectation is under \mbox{$\trajdist(\traj) = \trajdist(\state_1)\prod_{t=1}^T \trajdist(\state_{t+1}|\st,\at)\trajdist(\at|\st)$}, where $\trajdist(\state_{t+1}|\st,\at)$ is the dynamics distribution and $\trajdist(\at|\st)$ is the controller that we would like to learn.

\begin{algorithm}[H]
{\small
\caption{RL with linear-Gaussian controllers}
\label{alg:trajopt}
\begin{algorithmic}[1]
\STATE initialize $\trajdist(\at|\st)$
\FOR{iteration $k = 1$ to $K$}
\STATE run $\trajdist(\at|\st)$ to collect trajectory samples $\{\traj_i\}$
\STATE fit dynamics $\trajdist(\state_{t+1}|\st,\at)$ to $\{\traj_j\}$ using linear regression with GMM prior
\STATE fit $\trajdist = \arg\min_{\trajdist} E_{\trajdist(\traj)}[\cost(\traj)] \text{ s.t. } \kl(\trajdist(\traj)\|\bar{\trajdist}(\traj)) \leq \epsilon$
\ENDFOR
\end{algorithmic}
}
\end{algorithm}
\vspace{-0.34cm}

The controller can be optimized using a variety of model-based and model-free methods \cite{dnp-spsr-13}. We employ a trajectory-centric algorithm that optimizes time-varying linear-Gaussian controllers, which can be thought of as a trajectory with time-varying linear stabilization \cite{la-lnnpg-14}. While linear-Gaussian controllers are simple, they admit a very efficient optimization procedure that works well even under unknown dynamics. This method is summarized in Algorithm~\ref{alg:trajopt}. At each iteration, we run the current controller $\trajdist(\at|\st)$ on the robot to gather $N$ samples (\mbox{$N = 5$} in all of our experiments), then use these samples to fit time-varying linear-Gaussian dynamics of the form \mbox{$\trajdist(\state_{t+1}|\st,\at) = \gauss(\fxt \st + \fut \at + \fct, \noise_t)$}. This is done by using linear regression with a Gaussian mixture model prior, which makes it feasible to fit the dynamics even when the number of samples is much lower than the dimensionality of the system \cite{la-lnnpg-14}. We also compute a second order expansion of the cost function around each of the samples, and average the expansions together to obtain a local approximate cost function of the form
\[
\cost(\st,\at) \approx \frac{1}{2}[\st;\at]\tr\costhesst[\st;\at] + [\st;\at]\tr\costgradt + \text{const}.
\]
When the cost function is quadratic and the dynamics are linear-Gaussian, the optimal time-varying linear-Gaussian controller of the form \mbox{$\trajdist(\at|\st) = \gauss(\Kpol_t\st+\kpol_t,\ucovart)$} can be obtained by using the LQR method. This type of iterative approach can be thought of as a variant of iterative LQR \cite{lt-ilqr-04}, where the dynamics are fitted to data. In our implementation, we initialize $\trajdist(\at|\st)$ to be a random controller that is centered around the initial state with low-gain PD feedback, in order to avoid unsafe configurations on the real robot.

One crucial ingredient for making this approach work well is to limit the change in the controller $\trajdist(\at|\st)$ at each iteration, since the standard iterative LQR approach can produce a controller that is arbitrarily far away from the previous one, and visits parts of the state space where the fitted dynamics are no longer valid. To that end, our method solves the following optimization problem at each iteration:
\[
\min_{\trajdist(\at|\st)} E_{\trajdist(\traj)}[\cost(\traj)] \text{ s.t. } \kl(\trajdist(\traj)\|\bar{\trajdist}(\traj)) \leq \epsilon,
\]
\noindent where $\bar{\trajdist}(\traj)$ is the trajectory distribution induced by the previous controller. Using KL-divergence constraints for controller optimization was proposed in a number of prior works \cite{bs-cps-03,ps-rlmsp-08,pma-reps-10}. In the case of linear-Gaussian controllers, we can use a modified LQR algorithm to solve this problem. We refer the reader to previous work for details \cite{la-lnnpg-14}.


\subsection{Learning Nonlinear Policies with Guided Policy Search}

Linear-Gaussian controllers are easy to train, but they cannot express all possible control strategies, since they essentially encode a trajectory-following controller. To extend the applicability of linear-Gaussian controllers to more complex tasks, prior work has proposed to combine it with guided policy search \cite{la-lnnpg-14,lwa-lnnpg-15,levine2015end}, which is an algorithm that trains more complex policies, such as deep neural networks, by using supervised learning. The supervision is generated by running a simpler algorithm, such as the one in the previous section, from multiple initial states, generating multiple solutions. In our experimental evaluation, we combine linear-Gaussian controllers with guided policy search to learn a policy for hanging a loop of rope on a hook at different positions. A different linear-Gaussian controller is trained for each position, and then a single neural network policy is trained to unify these controllers into a single policy that can generalize to new hook positions. A full description of guided policy search is outside the scope of this paper, and we refer the reader to previous work for details \cite{la-lnnpg-14,lwa-lnnpg-15,levine2015end}.

\section{Algorithm Overview}
\label{sec:overview}

Our algorithm consists of three stages. In the first stage, we optimize an initial controller for the task without using vision, using the method described in the previous section. The state space during this stage corresponds to the joint angles and end-effector positions of the robot, as well as their time derivatives. This controller is rarely able to succeed at tasks that require keeping track of objects in the world, but it serves to produce a more focused exploration strategy that can be used to collect visual data, since an entirely random controller is unlikely to explore interesting parts of the state space. In general, this stage can be replaced by any reasonable exploration strategy.

\begin{figure}
\vspace{-0.08cm}
\setlength{\unitlength}{0.5\columnwidth}
\tikzstyle{block} = [rectangle, draw, 
    text width=6em, fill=white, align=center, rounded corners, node distance=2.73cm]
\tikzstyle{line} = [draw, -latex']
\pgfdeclarelayer{background layer} 
\pgfdeclarelayer{foreground layer} 
\pgfsetlayers{background layer,main,foreground layer} 
\definecolor{bgcolor}{RGB}{133, 224, 255}
\begin{tikzpicture}[node distance = 2cm]
    \node [block, text width=7.5em,  text depth=1.8em] (simple) {\footnotesize\vspace*{-2pt} \begin{spacing}{1}Train controller using $\tst$; collect image data\end{spacing}\vspace*{0pt}};
    \node [block, right of=simple, text width=4.2em, text depth=1.8em] (feat) {\footnotesize \vspace*{-2pt} \begin{spacing}{1}Learn visual features\end{spacing}\vspace*{0pt}};
    \node [block, right of=feat, text width=7.3em,  text depth=1.8em] (final) {\footnotesize \vspace*{-2pt} \begin{spacing}{1}  \hspace*{-6pt} Train final controller \\ \hspace*{-1pt}with new state $[\tst;\feat_t]$\hspace*{-4pt}\end{spacing}\vspace*{0pt}};
    \node [below of=feat, yshift=10pt,xshift=-45pt] (pic1) {\includegraphics[scale=0.28]{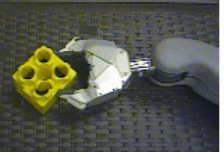}};
    \node [below of=feat, yshift=10pt,xshift=+45pt] (pic2) {\includegraphics[scale=0.28]{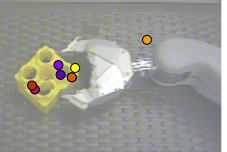}};
    \begin{pgfonlayer}{background layer}
	    \node [draw=black!50,  fill=bgcolor, rounded corners, fit={(simple) (feat) (final)}] {};
	    \end{pgfonlayer}
    \path [line] (simple) -- (feat);
    \path [line] (feat) -- (final);
    \path [line] (pic1) -- (pic2);
    \end{tikzpicture}
\caption{RL with deep spatial autoencoders. We begin by training a controller without vision to collect images (left) that are used to learn a visual representation (center), which is then used to learn a controller with vision (right). The learned state representation corresponds to spatial feature points (bottom).
\label{fig:diagram}
}
\vspace{-0.2in}
\end{figure}
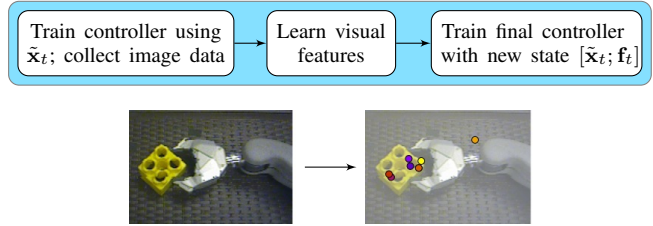

In the second stage, the initial controller is used to collect a dataset of images, and these images are then used to train our deep spatial autoencoder with unsupervised learning, as described in Section~\ref{sec:ae}. Once this autoencoder is trained, the encoder portion of the network can be used to produce a vector of feature points for each image that concisely describes the configuration of objects in the scene. The final state space is formed by concatenating the joint angles, end-effector positions, and feature points, and also including their velocities to allow for dynamic tasks to be learned. We must also define a cost function using this new state space, which we do by presenting an image of the target state to the feature encoder and determining the corresponding state.

In the third stage, a vision-based controller is trained using the new state space that now contains visual features from the encoder, with the new cost function defined in terms of the visual features and the robot's configuration. This controller is trained using the same trajectory-centric reinforcement learning algorithm as in the first stage, and is able to perform tasks that require controlling objects in the world that are identified and localized using vision.

An overview of this procedure is provided in Figure~\ref{fig:diagram}. In the next section, we describe our representation learning method.


\section{Unsupervised State Representation Learning from Visual Perception}
\label{sec:ae}

The goal of state representation learning is to find a mapping $\autoenc(\img_t)$ from a high-dimensional observation $\img_t$ to a robot state representation for which it is easy to learn a control policy. We will use $\tst$ to denote the configuration of the robot, $\feat_t$ to denote the learned representation, and $\st = [\tst;\feat_t]$ to denote the final state space that combines both components. The state of a robotic system should be consistent with properties of the physical world~\cite{jb-srl-14}; i.e., it should be temporally coherent and encode only simple, task-relevant information. Moreover, our RL method models the system's state as a time-varying linear dynamical system, and thus it will perform best if the state representation acts as such. Intuitively, a good state representation should encode the poses of relevant objects in the world using Cartesian coordinates that move with the object over time. Our aim is to learn such a state representation without using human supervision or hand-crafted features. Deep learning offers powerful tools for learning representations \cite{lbh-dl-15}, but most deep neural network models focus on semantics of \emph{what} is in the scene,
rather than \emph{where} the objects are. In this section, we describe our autoencoder architecture which is designed to encode temporally-smooth, spatial information, with particular emphasis on object locations. As we show in our experiments, this kind of representation is well suited to act as a state space for robotic reinforcement learning.

\subsection{Deep Spatial Autoencoders}
\label{sec:dsae}

\begin{figure*}[t]
\centering
\setlength{\unitlength}{1.0\columnwidth}
\begin{picture}(1.99,0.3) \linethickness{0.5pt}
\put(0,0.00){\includegraphics[width=\textwidth]{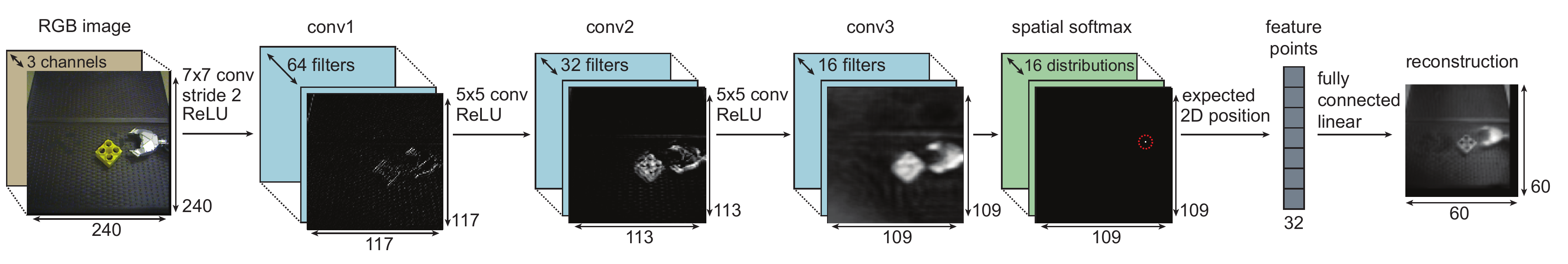}}
\put(0.16,0.3){\footnotesize{$I$}}
\put(1.875,0.254){\footnotesize{$\autodec(\feat)$}}
\end{picture}
\vspace{-0.1in}
\caption{The architecture for our deep spatial autoencoder. The last convolutional layer is passed through a spatial softmax, followed by an expectation operator that extracts the positions of the points of maximal activation in each channel. A downsampled version of the image is then reconstructed from these feature points.
}
\vspace{-0.15in}
\label{fig:ae}
\end{figure*}

Autoencoders acquire features by learning to map their input to itself, with some mechanism to prevent trivial solutions, such as the identity function. These mechanisms might include sparsity or a low-dimensional bottleneck. Our autoencoder architecture, shown in Figure~\ref{fig:ae}, maps from full-resolution RGB images to a down-sampled, grayscale version of the input image, and we force all information in the image to pass through a bottleneck of spatial features, which we describe below. Since low-dimensional, dense vectors are generally well-suited for control, a low-dimensional bottleneck is a natural choice for our learned feature representation. A critical distinction in our approach is to modify this bottleneck so that it is forced to encode spatial feature locations rather than feature values -- the ``where'' rather than the ``what.'' This architecture makes sense for a robot state representation, as it forces the network to learn object positions; we show in Section~\ref{sec:results}, that it outperforms more standard architectures that focus on the ``what,'' both for image reconstruction, and for robotic control. The final state space for RL is then formed by concatenating this learned encoding, as well as its time derivatives (the feature ``velocities''), with the robot's configuration.

The first part of the encoder architecture is a standard 3-layer convolutional neural network with rectified linear units of the form $\responsepix=max(0,z_{cij})$ for each channel $\channel$ and pixel $(i,j)$. We compute the spatial features from the last convolutional layer by performing a ``spatial soft arg-max'' operation that determines the image-space point of maximal activation in each channel of conv3. This set of maximal activation points forms our spatial feature representation and forces the autoencoder to focus on object positions. The spatial soft arg-max consists of two operations. The response maps of the third convolutional layer (conv3) are first passed through a spatial softmax \mbox{$\softmaxpix = {e^{\responsepix/\temp}}/{\sum_{i' j'}{e^{\responsepixprime/\temp}}}$}, where the temperature $\alpha$ is a learned parameter. Then, the expected 2D position of each softmax probability distribution $\softmax_c$ is computed according to \mbox{$\feat_c = (\sum_i i * \softmaxpix ,  \sum_j j * \softmaxpix)$}, which forms the bottleneck of the autoencoder. This pair of operations typically outputs the image-space positions of the points of maximal activation for each channel of conv3. Finally, the decoder $\autodec$ is simply a single linear (fully connected) mapping from the feature points $\feat$ to the down-sampled image. We found that this simple decoder was sufficient to acquire a suitable feature representation.

With this architecture, the bottleneck representation $\feat = h_\text{enc}(I)$, which we refer to as learned ``feature points,'' encodes the positions of the learned features in the image. The autoencoder is forced to compress all of the image information through this spatial feature point representation, which will therefore be capable of directly localizing objects in the image. This means that the state representation used for control will capture the spatial organization of objects in the scene. Example learned feature points are visualized in Figure~\ref{fig:fppresence}. Note that these feature points pick out task-relevant objects (those which move during the collection of the image data.), and learn to ignore background features. One drawback of this representation, however, is that there is always exactly one output point per feature channel, which does not gracefully handle occlusions (zero features present) or repeated structure and objects (more than one feature).

A feature representation that can reconstruct the image clearly contains the state information captured by the camera. However, the learning algorithm must also be able to predict the dynamics of the state representation, which is not necessarily inherently possible with unconstrained learned features~\cite{jb-srl-14}. Thus, we add a penalty to the autoencoder objective, \mbox{$g_{\operatorname{slow}}(\feat_t) = ||(\feat_{t+1}-\feat_{t}) - (\feat_{t}-\feat_{t-1})||_2^2$}, to encourage the feature points to slowly change velocity. As a result, the overall autoencoder objective becomes:
$$\mathcal{L}_{\text{DSAE}} = \sum_{t,k}||I_{\operatorname{downsamp},k,t}-\autodec(\feat_{k,t}) ||_2^2 + g_{\operatorname{slow}}(\feat_{k,t})
$$
where $I_{\operatorname{downsamp}}$ is a downsampled, grayscale version of the input image $I$, and $\feat_{k,t} =h_\text{enc}(I_{k,t})$, the feature point encoding of the $t$th image in the $k$th sequence.

We optimize the auto encoder using stochastic gradient descent (SGD), and with batch normalization~\cite{is-bn-15} following each of the convolutional operations. The filters of the first convolutional layer are initialized with a network trained on ImageNet~\cite{ddsll-lshid-09,sljsr-gdwc-14}.


\subsection{Filtering and Feature Pruning Using Feature Presence}
\label{sec:featselect}

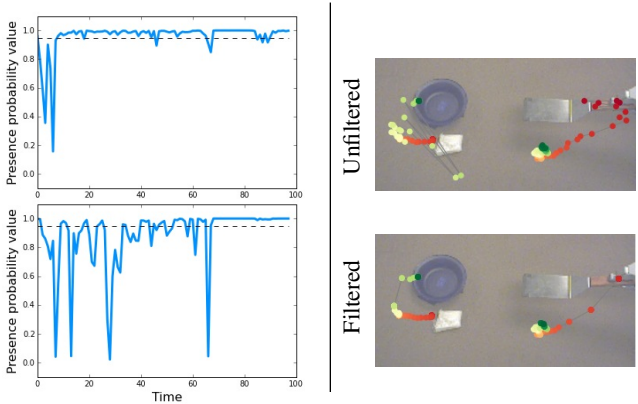
\begin{figure}
\setlength{\unitlength}{0.5\columnwidth}
\begin{picture}(1.99,1.2) \linethickness{0.5pt}

\put(0.04,0.58){\includegraphics[width=0.47\columnwidth]{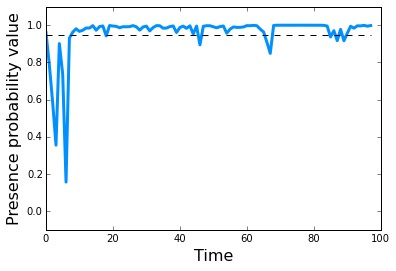}}
\put(0.04,0.00){\includegraphics[width=0.47\columnwidth]{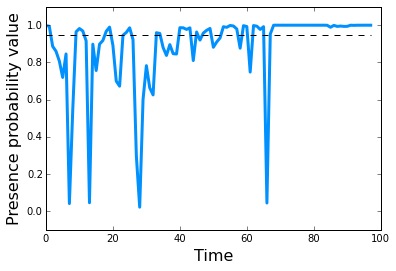}}

\put(1.19,0.67){\includegraphics[width=0.4\columnwidth]{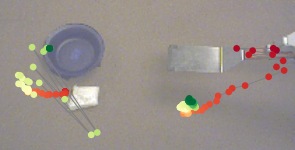}}
\put(1.19,0.13){\includegraphics[width=0.4\columnwidth]{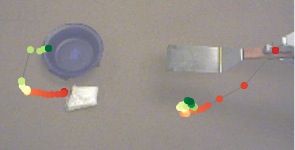}}

\put(1.09,0.72){\rotatebox{90}{\small Unfiltered}}
\put(1.09,0.23){\rotatebox{90}{\small Filtered}}

\linethickness{0.5pt}
\put(1.05,0.05){\line(0,1){1.2}}

\end{picture}
\vspace{-0.55cm}
\caption{Left: feature presence probabilities plotted for two features, with the threshold $\beta$ shown in black. Note that a range of thresholds would produce similar results. Right: two sample trajectories of learned feature points, starting from red and ending in green. The features filtered with a Kalman filter using feature presence (bottom) are much smoother than the unfiltered feature points (top)}
\label{fig:fppresence}
\vspace{-0.2in}
\end{figure}
Not all of the learned feature points will encode useful information that can be adequately modeled with a time-varying linear dynamical system by our RL algorithm. Lighting and camera exposure can cause sudden changes in the image that do not reflect motion of relevant objects. However, the autoencoder will assign some subset of its features to model these phenomena to improve its reconstruction, producing noisy feature points that make it difficult to train the controller. Additionally, even useful features can periodically become occluded, resulting in low activations in the last convolutional layer and unpredictable peak locations.

To handle these issues, we introduce the concept of feature presence. The presence of a feature is determined by the softmax probability value $\softmax_{cij}$ at the pixel location of the feature point $\feat_c = (i,j)$.\footnote{Since the feature point position may fall on a boundary between pixels, we actually sum the probability over a $3 \times 3$ window of nearby pixels.} This measure reflects the probability mass after the softmax that is actually located at the position of the maximum activation. This measure is usually very high. As shown in Figure~\ref{fig:fppresence}, learned feature points usually place almost all of their probability mass at $\feat_c$, the location of the maximum. We therefore use a threshold on $\softmax_{cij}$ that marks the feature as present if $\softmax_{cij} \geq \beta$, with $\beta = 0.95$. The positions of the feature points during a trial are filtered by a Kalman filter, which receives observations of $\feat_c$ only when $\softmax_{cij} \geq \beta$, producing smooth predictions when the feature is not present. We use a second order Kalman filter and fit the parameters using expectation maximization.

In addition to filtering features that might become occluded, we also automatically determine the features that do not adequately represent task-relevant objects in the scene and prune them from the state. As discussed in Section~\ref{sec:overview}, the goal positions for the feature points are defined by showing an image of the goal state to the robot. To mitigate the effects of noise, we record a short sequence of fifty images over the course of two seconds and average the resulting feature positions. We also measure the feature presence indicators during this period, and prune those points that are not present for the entire sequence, since they are unlikely to be relevant. In our evaluation, we show that this pruning step is important for removing extraneous and noisy features from the state. To validate this heuristic, we conducted a more thorough analysis of the predictability of the feature points to determine its effectiveness

To analyze the usefulness of the learned feature points for modeling the dynamics of the scene, we define the following measure of feature \emph{predictiveness}:
\begin{equation*}
E\!\left[\sum_{t=1}^T \left(\log p(\state_t^\featidxsel \mid \state^\featidxsel_{t-1}, \action_{t-1}) \!+\! \log p(\feat_t^{\featidx \setminus \featidxsel} \mid \feat_t^\featidxsel)\right) \right]
\end{equation*}
Here, $\featidx$ denotes the index set of all features, $\featidxsel \subset \featidx$ denotes the indices that are actually selected for inclusion in the state, $\feat^\featidxsel$ is shorthand for the corresponding feature vectors, and $\state^\featidxsel = [\tst; \feat^\featidxsel]$ is the corresponding state space that includes the robot's configuration. We evaluate the probabilities by fitting a dynamics model to $\state^\featidxsel$ as described in Section~\ref{sec:rl} and fitting a linear-Gaussian model to $p(\feat_t^{\featidx \setminus \featidxsel} \mid \feat_t^\featidxsel)$. The expectation is evaluated at samples gathered for autoencoder training. This measure captures the degree to which the state $\state^\featidxsel_{t-1}$ is predictive of the next state $\state^\featidxsel_t$, as well as the degree to which $\state^\featidxsel_t$ can predict $\feat_t^\featidx$. Intuitively, a good state space can predict the features at the next step, which are optimized for predicting the entire image. This measure will discourage features with dynamics that are hard to fit, and will also encourage diversity, since multiple features on the same object will be predictable from one another according to $p(\feat_t^{\featidx \setminus \featidxsel} \mid \feat_t^\featidxsel)$. To analyze the relative quality of each feature, we iteratively pruned out the feature that, when removed, produced the least reduction in the likelihood. This yields a predictiveness ranking of the features. We evaluated this metric on the most difficult task, and found that our simple feature presence metric was a good, quick approximation for selecting highly-predictive features. We report the rankings, along with the chosen features in Table~\ref{tbl:featrank}. We also found that the low-ranked features are always either consistently not present or extremely noisy (see Figure~\ref{fig:fppresence}).

\begin{table}[h!]
\vspace{-0.1cm}
{\footnotesize
  \begin{center}
    \begin{tabular}{| l | c | c | c | c |}
    \hline
    Task & \specialcell{feature ranking (from best to worst)} \\
    \hline
    Rice scooping &  \textbf{5} \textbf{15} 12 \textbf{1}  4  6  2  14  8  13  7  3  10  9  11  16\\
    \hline
    \end{tabular}
  \end{center}
}
\vspace{-0.1in}
  \caption{Ranking of features according to predictiveness; features chosen based on presence in the target image are shown in bold.}
  \label{tbl:featrank}
\vspace{-0.15in}

\end{table}

\section{Control Over Visual Features}

\begin{algorithm}[b]
{\small
\caption{RL with deep spatial autoencoders}
\label{alg:summary}
\begin{algorithmic}[1]
\STATE train exploration controller $\tilde{\trajdist}(\at|\tst)$ with simple states $\tst$ and simple cost function $\cost(\tst,\at)$ (Algorithm~\ref{alg:trajopt})
\STATE collect image dataset $\mathcal{D} = \{\img_k\}$ by running $\tilde{\trajdist}(\at|\tst)$ (or using images collected during step 1)
\STATE train deep spatial autoencoder using $\mathcal{D}$ to obtain feature encoder $\autoenc(\img_t) = \feat_t$
\STATE select feature points $\tilde{\feat}$ using heuristic described in Section~\ref{sec:featselect}, define full state as $\st = [\tst;\tilde{\feat_t};\dot{\tilde{\feat_t}}]$ 
\STATE define full cost function $\cost(\st,\at)$ using images of the target state
\STATE train final controller $\trajdist(\at|\st)$ with full cost $\cost(\st,\at)$ (Algorithm~\ref{alg:trajopt})
\end{algorithmic}
}
\end{algorithm}
\vspace{-0.1cm} 

Putting everything together, our method for training vision-based controllers is outlined in Algorithm~\ref{alg:summary}. We first train a controller to attempt the task without vision, using only the robot's configuration as the state. The goal for this controller is defined in terms of the robot's configuration, and the learning is performed using Algorithm~\ref{alg:trajopt}. While this goal is typically inadequate for completing the task, it serves to explore the state space and produce a sufficiently varied range of images for learning a visual state representation. The representation is then learned on a dataset of images obtained from this ``blind'' controller using the method described in Section~\ref{sec:dsae}. This produces a feature encoder $\autoenc(\img_t) = \feat_t$. We then train Kalman filters for each feature point and automatically prune the features using the heuristic described in Section~\ref{sec:featselect}, yielding the final features $\tilde{\feat}$ to be included in the full state $\st = [\tst;\tilde{\feat_t};\dot{\tilde{\feat_t}}]$. Since we use torque control, the robot and its environment form a second-order dynamical system, and we must include both the features $\tilde{\feat}$ and their velocities $\dot{\tilde{\feat}}$ in the state.

In addition to the state representation $\st$, we must also define a cost function $\cost(\st,\at)$ for the task. As discussed in the preceding section, we set the goal by showing the robot an image of the desired configuration of the scene, denote $\img_\text{goal}$, in addition to the target end-effector position used for the ``blind'' controller. The target features are then obtained according to $\feat_\text{goal} = \autoenc(\img_\text{goal})$ and the cost depends on the distance to these target features. In practice, we use a short sequence of images to filter out noise. Having defined a state space $\st$ and a cost $\cost(\st,\at)$, we use Algorithm~\ref{alg:trajopt} to optimize the final vision-based controllers that perform feedback control on the robot's configuration and the visual state representation.

\section{Experimental Evaluation}
\label{sec:results}

We evaluated our method on a range of robotic manipulation tasks, ranging from pushing a lego block to scooping a bag of rice into a bowl. The aim of these experiments was to determine whether our method could learn behaviors that required tracking objects in the world that could only be perceived with vision. To that end, we compared controllers learned by our approach to controllers that did not use vision, instead optimizing for a goal end-effector position. We also compared representations learned with our spatial autoencoder to hidden state representations learned by alternative architectures, including ones proposed in previous work.

\subsection{Experimental Tasks}

\begin{figure}
\setlength{\unitlength}{0.5\columnwidth}
\begin{picture}(1.99,1.02) \linethickness{0.5pt}

\put(0.08,0.56){\includegraphics[width=0.23\columnwidth]{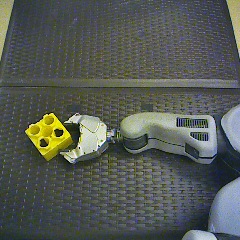}}
\put(0.08,0.08){\includegraphics[width=0.23\columnwidth]{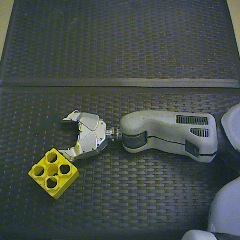}}

\put(0.56,0.56){\includegraphics[width=0.23\columnwidth]{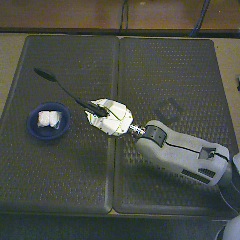}}
\put(0.56,0.08){\includegraphics[width=0.23\columnwidth]{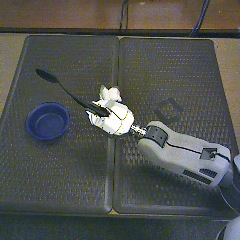}}

\put(1.04,0.56){\includegraphics[width=0.23\columnwidth]{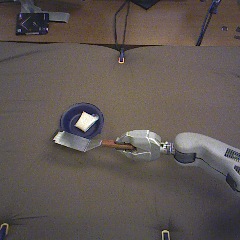}}
\put(1.04,0.08){\includegraphics[width=0.23\columnwidth]{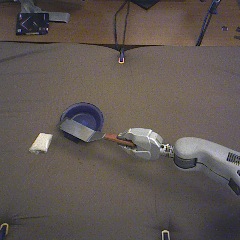}}

\put(1.52,0.56){\includegraphics[width=0.23\columnwidth]{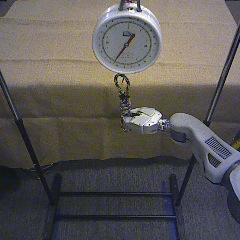}}
\put(1.52,0.08){\includegraphics[width=0.23\columnwidth]{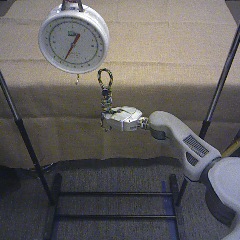}}

\newlength{\illuwidth}
\setlength{\illuwidth}{0.23\columnwidth-6pt}
\put(0.08,0){\colorbox{white}{\makebox[\illuwidth]{\footnotesize lego block}}}
\put(0.56,0){\colorbox{white}{\makebox[\illuwidth]{\footnotesize bag transfer}}}
\put(1.04,0){\colorbox{white}{\makebox[\illuwidth]{\footnotesize rice scoop}}}
\put(1.52,0){\colorbox{white}{\makebox[\illuwidth]{\footnotesize loop hook}}}

\put(0.0,0.69){\rotatebox{90}{success}}
\put(0.0,0.2){\rotatebox{90}{failure}}

\end{picture}
\caption{Illustrations of the tasks in our experiments, as viewed by the robot's camera, showing examples of successful completion by our controllers and a typical failure by the controller without vision. 
}
\label{fig:tasks}
\vspace{-0.19in}
\end{figure}
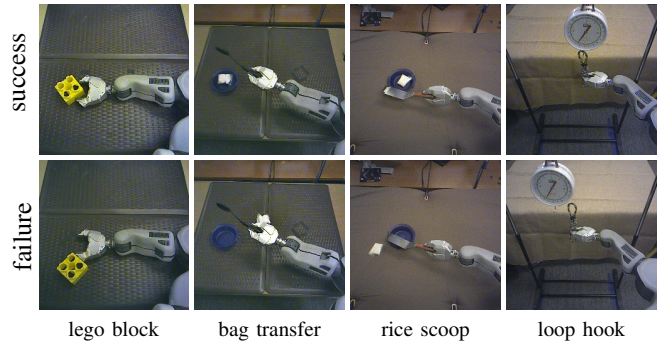

The four experimental tasks in our evaluation are shown in Figure~\ref{fig:tasks}. The first and simplest task requires sliding a lego block 30 cm across a table. The robot does not grasp the lego block, but rather pushes it along the table with its gripper, which requires coordinating the motion of the block and balancing it at the end of the fingers. In the second task, the robot is holding a spoon that contains a white bag, and it must use the spoon to drop the bag into a bowl. The robot must balance the bag on the spoon without dropping it, and must angle the spoon such that the bag falls into the bowl. In the third task, the robot must use a spatula to scoop a bag of rice off of the table and into a bowl. In this task, the robot must perform a quick, precise sliding motion under the bag and then lift and angle the spatula such that the rice bag slides or flips into the bowl. This task is illustrated in detail in Figure~\ref{fig:teaser}. If the robot does not coordinate the motion of the spatula and rice bag, the bag is pushed aside and cannot be lifted properly.

In the fourth task, we combine our method with guided policy search by training four separate linear-Gaussian controllers for hanging a loop of rope on a hook at different positions. These four controllers are combined into a single neural network policy using guided policy search, and we test the capability of this policy to then hang the rope on the hook in new positions. The robot must use vision to locate the hook, so making use of the spatial feature points is essential for performing this task.

All of our experiments were performed using one 7 DoF arm of a PR2 robot. The initial state space $\tst$ contains the robot's joint angles and end effector pose, as well as their time derivatives. The controls consist of the seven torques at the joints. The images were collected using a consumer RGB camera, and the controller runs at 20 Hz. For each of the tasks, we provided the goal pose of the end-effector and an image of the goal state. For the scooping task, we also provided a sub-goal of the spatula under the bag of rice.
Full details of each task and the equation for the cost function are both presented in Appendix~\ref{app:tasks} of the supplementary materials.\footnote{All supplementary materials and videos of the learned controllers can be viewed on the project website: {\footnotesize \url{http://rll.berkeley.edu/dsae}}.}

\subsection{Results and Analysis}

We ran each of the learned controller 10 times and reported either the average distance to the goal, or the success rate. For the transfer and scooping task, success required placing the bag inside the bowl, while for the hanging task, success required the rope loop to hang on the hook. For each of the tasks, we compared our method to a controller that used only the robot's configuration $\tst$. The results, shown in Table~\ref{tbl:novision}, show that each of the tasks requires processing visual information, and that our method is able to acquire a state that contains the information necessary for the task.


\vspace{-0.2cm}
\begin{table}[h!]
{\footnotesize
  \begin{center}
    \begin{tabular}{| l | c | c | c | c |}
    \hline
    task & \specialcell{lego block\\(mean distance)} & \specialcell{bag transfer} & rice scoop & loop hook \\
    \hline
    ours & 0.9 cm & 10/10 & 9/10 & 60/70 \\
    \hline
        \hline
    no vision &  4.6 cm & 1/10 & 0/10 & 7/70 \\
    \hline
    \end{tabular}
  \end{center}
}
\vspace{-0.1in}
  \caption{Results of our method. Controllers trained without vision fail to complete the tasks, illustrating that the tasks require vision.
 }
  \label{tbl:novision}
\vspace{-0.1in}
\end{table}

For the generalization experiment with the rope loop on hook task, we trained linear-Gaussian controllers for hook positions that were spaced 8 cm apart, and trained a single neural network using guided policy search that unified these controllers into a single nonlinear policy. This nonlinear policy could succeed at each of the training positions, and was also successful at the three test positions not seen during training, as shown in Table~\ref{tbl:gen}. The policy trained without vision was unable to localize the hook, and resorted to a random strategy, which was occasionally successful but generally failed at the task. The supplementary video shows the behavior of both the ``blind'' policy and the policy that used our learned visual state representation.


\newcommand{\incm}[1]{\hspace{-2pt}\footnotesize #1 cm\hspace{-2pt}}
\newcommand{\sq}[1]{\hspace{-2pt}\footnotesize #1{\hspace{-2pt}}}
\vspace{-0.2cm}
\begin{table}[h!]
\begin{center}
\begin{tabular}{| l | c | c | c | c | c | c | c |}
\hline

\multicolumn{1}{|l}{\footnotesize loop hook} & \multicolumn{4}{|c}{\footnotesize training positions} & \multicolumn{3}{|c|}{\footnotesize test positions}\\
\hline
{\footnotesize hook position} & \incm{0} & \incm{8} & \incm{16} & \incm{24} & \incm{4} & \incm{12} & \incm{20} \\
\hline
{\footnotesize ours} &  \sq{10/10} &  \sq{8/10} &  \sq{9/10} &  \sq{8/10} &  \sq{8/10} &  \sq{10/10} &  \sq{7/10} \\
\hline
\hline
{\footnotesize no vision} & \sq{0/10} &  \sq{1/10} &  \sq{0/10} &  \sq{3/10} &  \sq{2/10} &  \sq{0/10} &  \sq{1/10} \\
\hline
\end{tabular}
\end{center}
\vspace{-0.1in}
\caption{Detailed generalization results for the rope loop task.
\label{tbl:gen}
}
\vspace{-0.1in}
\end{table}

We also evaluate two alternative autoencoder architectures that are representative of prior work, trained with the same training and validation datasets as our model. The first comparison closely mirrors the method of Lange et al.~\cite{rlv-arlrv-12}, but with a bottleneck dimensionality of $10$ to account for the greater complexity of our system. The second comparison reflects a more recent architectures, using max-pooling to reduce the dimensionality of the image maps and batch normalization after the convolutional layers. The bottleneck for this architecture is $32$, matching that of our architecture.
Details of both baseline architectures are provided in Appendix~\ref{app:baseline} of the supplementary materials$^2$.
We evaluated both architectures with and without a smoothness penalty. The results, shown in Table~\ref{tbl:baselines}, show that these methods struggle with our high-dimensional, real-world tasks, despite the larger model achieving a lower reconstruction loss than our autoencoder. For several of the models, we could not obtain a stable controller, while the others did not effectively integrate visual input into the control strategy. Table~\ref{tbl:baselines} also shows the performance of our method without the smoothness penalty, and without feature pruning, showing that both of these components are critical for obtaining good results.



\vspace{-0.2cm}
\begin{table}[h!]
{\footnotesize
  \begin{center}
    \begin{tabular}{| l | r | r | r |}
    \hline
    \!\!lego block comparison \!\!\!& \!\!\specialcell{reconstruction\\loss (validation)}\!\! & \!\!\specialcell{network\\training time}\!\! & \!\!\specialcell{mean\\distance (cm)}\!\!\\
    \hline
    \!\!AE of~\cite{rlv-arlrv-12} & 7.23 & 270 min & 8.4 $\pm$ 6.5 \\  
    \hline
    \!\!AE of \cite{rlv-arlrv-12}, smooth & 7.43 & 420 min & 9.7 $\pm$ 8.0  \\
    \hline
    \!\!conv+pool AE & 5.29 & 11 min & \!\!\! \textsuperscript{*}24.3 $\pm$ 6.0  \\
    \hline
    \!\!conv+pool AE, smooth \!\!\!\!\!& 5.33 & 32 min & \!\!\! \textsuperscript{*}30.0 $\pm$ 0.0 \\
    \hline
    \!\!ours, no smooth & 6.07 & 27 min & 5.1 $\pm$ 1.5   \\
    \hline
\!\!ours, no feat pruning & 6.01 & 80 min & \!\!\! \textsuperscript{*}30.0 $\pm$ 0.0 \\
    \hline
    \!\!ours & 6.01 & 80 min & {\bf 0.9 $\pm$ 0.3}   \\
    \hline
    \hline
    \!\!No vision & n/a & 0 min & 4.6 $\pm$ 3.3  \\
    \hline
    \multicolumn{4}{l}{\textsuperscript{*}\footnotesize{These controllers became unstable and did not train successfully.}}\\
    \end{tabular}
  \end{center}
}
\vspace{-0.1in}
  \caption{Comparisons to prior autoencoder architectures and variants of our method.}
  \label{tbl:baselines}
\vspace{-0.1in}
\end{table}

Another advantage of our approach is its sample efficiency, which is enabled both by the use of simple linear-Gaussian controllers and a data-efficient neural network architecture. The autoencoders used around 50 trials for each task, with each trial consisting of 100 image frames and 5 second of interaction time, for a total of 5000 frames per task. Training the final vision-based controller required another 50-75 trials, which means that each controller was trained with around 10-15 minutes of total robot interaction time.

\section{Discussion and Future Work}

We presented a method for learning state representations using deep spatial autoencoders, and we showed how this method could be used to learn a range of manipulation skills that require close coordination between perception and action. Our method uses a spatial feature representation of the environment, which is learned as a bottleneck layer in an autoencoder. This allows us to learn a compact state from high-dimensional real-world images. Furthermore, since this representation corresponds to image-space coordinates of objects in the scene, it is particularly well suited for continuous control. The trajectory-centric RL algorithm we employ can learn a variety of manipulation skills with these spatial representations using only tens of trials on the real robot.

While we found that controllers trained without vision could adequately explore the space for each task to generate training data for representation learning, there are tasks where visiting a sufficient range of states can be difficult without vision. A promising direction for tackling such tasks is to interleave representation learning with controller optimization in a similar spirit to iterative model learning approaches \cite{rb-asimb-12}. Another promising future direction is to use additional sensory modalities to learn more advanced sensory state representations, e.g. depth and haptic sensing.

The prospect of autonomously learning compact and portable state representations of complex environments entirely from raw sensory inputs, such as images, has widespread implications for autonomous robots. The method presented in this paper takes one step in this direction, by demonstrating that a spatial feature architecture can effectively learn suitable representations for a range of manipulation tasks, and these representations can be used by a trajectory-centric reinforcement learning algorithm to learn those tasks. Further research in this direction has the potential to make it feasible to apply out-of-the-box learning-based methods to acquire complex skills for tasks where a state space is very difficult to design by hand, such as tasks with deformable objects, complex navigation in unknown environments, and numerous other tasks.






\noindent
\textbf{Acknowledgements:} {\small This research was funded in part by ONR through a Young Investigator Program award, by the Army Research Office through the MAST program, and by the Berkeley Vision and Learning Center (BVLC). Chelsea Finn was also supported in part by a SanDisk fellowship.}


\bibliographystyle{ieeetran}
\bibliography{references}
\newpage

\appendix
\subsection{Robotic Experiment Details}
\label{app:tasks}

All of the robotic experiments were conducted on a PR2 robot. The robot was controlled at 20 Hz via direct effort control,\footnote{The PR2 robot does not provide for closed loop torque control, but instead supports an effort control interface that directly sets feedforward motor voltages. In practice, these voltages are roughly proportional to feedforward torques, but are also affected by friction and damping.} and camera images were recorded using the RGB camera on a PrimeSense Carmine sensor. The images were downsamples to $240 \times 240$. The learned policies controlled one 7 DoF arm of the robot. The camera was kept fixed in each experiment. Each episode was $5$ seconds in length. For each task, the cost function required reaching the goal state, defined both by visual features and gripper pose.
Similar to previous work, the cost was given by the following equation:
\[
\cost(\st,\at) = w_{\ell_2} d_t ^ 2 + w_{\log} \log(d_t^2 + \alpha) + w_{\action}\vnorm{\at}^2,
\]
\noindent where $d_t$ is the distance between three points in the space of the end-effector and learned feature points in 2D and their respective target positions\footnote{Three points fully define the pose of the end-effector.}, and the weights are set to $w_{\ell_2} = 10^{-3}$, $w_{\log} = 1.0$, and $w_{\action} = 10^{-2}$. The quadratic term in the cost encourages moving towards the target when it is far, while the logarithm term encourages reaching the target state precisely, as discussed in prior work~\cite{lwa-lnnpg-15}. The rice scoop task used two target states, in sequence, with half of the episode ($2.5$ seconds) devoted to each target. For each of the tasks, the objects were reset to their starting positions manually between trials during training. We discuss the particular setup for each experiment below:

\paragraph{Lego block} The lego block task required the robot to push a lego block $30\text{ cm}$ to the left of its initial position. For this task, we measured and reported the distance between the top corner of the goal block position to the nearest corner of the lego block at the end of the trial. In some of the baseline evaluations, the lego block was flipped over, and the nearest corner was still used to measure distance to the goal.

\paragraph{Bag transfer} The bag transfer task required the robot to place a white bag into a bowl, using a spoon. At the start of each trial, the robot was grasping the spoon with the bag in the spoon. A trial was considered successful if the bag was inside the bowl and did not extend outside of the bowl. In practice, the bag was very clearly entirely in the bowl, or entirely outside of the bowl during all evaluations.

\paragraph{Rice scoop} The rice scooping task required the robot to use a spatula to lift a small bag of rice off of a table and place it in a bowl. At the start of each trial, the spatula was in the grasp of the robot gripper, and the bag of rice was on the table, about $3\text{ cm}$ from the bowl. As with the bag transfer task, a trial was considered successful if the bag of rice was inside the bowl and did not extend outside of the bowl. In practice, the bag was very clearly in the bowl, or outside of the bowl during all evaluations.

\paragraph{Loop hook} The loop hook task required the robot to place a loop of rope onto a metal hook attached to a scale, for different positions of the hook. At training time, the scale was placed at four different starting positions along a metal pole that were equally spaced across $24\text{ cm}$ of the pole. The test positions were the three midpoints between the four training positions. A trial was considered successful if, upon releasing the rope, the loop of rope would hang from the hook. In practice, the failed trials using our approach were often off by only $1$-$2 \text{ mm}$, whereas the controller with no vision was typically off by several centimeters.

\subsection{Neural Network Architectures for Prior Work Methods}
\label{app:baseline}

We compare our network with two common neural network architectures. The first baseline architecture is the one used by Lange et al.~\cite{rlv-arlrv-12}. The network is composed of 8 encoder layers and 8 decoder layers. To match the original architecture as closely as possible, we converted our $240\times 240$ RGB images into $60\times 60$ grayscale images before passing them through the network. The encoder starts with 3 convolution layers with filter size $7\times 7$, where the last convolution layer has stride $2$. The last convolution layer is followed by 6 fully connected layers, the size of which are $288$, $144$, $72$, $36$, $18$ and $10$ respectively. The last fully connected layer forms the bottleneck of the autoencoder. We chose $10$ as the dimension of the bottleneck, since the system has roughly $10$ degrees of freedom. The decoder consists of 6 mirrored fully connected layers followed by $3$ deconvolution layers, finally reconstructing the down sampled $60\times 60$ image. We used ReLU nonlinearities between each layer. Following~\cite{rlv-arlrv-12}, we pre-train each pair of the encoder-decoder layers for $4000$ iterations. Then, we perform fine tuning on the entire network until the validation error plateaus. 

We also experimented with a more widely adopted convolutional architecture. The $240 \times 240 \times 3$ image is directly passed to the network. This network starts with 3 convolutional layers. As in our network architecture, conv1 consists of $64$ $7\times 7$ filters with stride $2$, conv2 has $32$ $5 \times 5$ filters with stride $1$, and conv3 has $16$ $5 \times 5$ filters with stride $1$, each followed by batch normalization and ReLU nonlinearities. Unlike our architecture, this baseline architecture performs max-pooling after each convolution layer in order to decrease the dimensionality of the feature maps. The convolution layers are followed by two fully connected layers with $512$ and $32$ units respectively, the last of which forms the bottleneck of the network. These layers together form the encoder, and a mirroring architecture, consisting of fully connected layers and deconvolution layers, forms the decoder. We initialize the first convolution layer with weights trained on ImageNet, and train the network until validation error plateaus. 


\end{document}